\documentclass[acmtog, nonacm]{acmart}

\renewcommand\footnotetextcopyrightpermission[1]{}
\usepackage{multirow}
\usepackage{enumitem}

\AtBeginDocument{%
  }

\definecolor{amber}{rgb}{1.0, 0.49, 0.0}
\definecolor{dodgerblue}{RGB}{30, 144, 255}
\definecolor{violet}{RGB}{238,130,238}
\definecolor{my_green}{RGB}{113,173,71}
\definecolor{my_blue}{RGB}{44,115,182}

\newcommand{\reffig}[1]{\textcolor{black}{Fig.~\ref{fig:#1}}} 
\newcommand{\refsec}[1]{\textcolor{black}{Sec.~\ref{sec:#1}}}
\newcommand{\reftab}[1]{\textcolor{black}{Tab.~\ref{tab:#1}}}

\newcommand{\eg}[1]{\textcolor{black}{\textit{e.g.,~}}}
\newcommand{\ie}[1]{\textcolor{black}{\textit{i.e.,~}}}

\begin{document}

\title{Text-to-Vector Generation with Neural Path Representation}


\author{Peiying Zhang}
\affiliation{
 \institution{City University of Hong Kong}
  \city{Hong Kong}
  \country{China}
 }
\email{zhangpeiying17@gmail.com}

\author{Nanxuan Zhao}
\affiliation{
 \institution{Adobe Research}
  \city{San Jose}
  \country{USA}
 }
\email{nanxuanzhao@gmail.com}

\author{Jing Liao}
\authornote{Corresponding author}
\affiliation{
 \institution{City University of Hong Kong}
 \city{Hong Kong}
 \country{China}
 }
\email{jingliao@cityu.edu.hk}



\begin{abstract}

Vector graphics are widely used in digital art and highly favored by designers due to their scalability and layer-wise properties. However, the process of creating and editing vector graphics requires creativity and design expertise, making it a time-consuming task. Recent advancements in text-to-vector (T2V) generation have aimed to make this process more accessible. However, existing T2V methods directly optimize control points of vector graphics paths, often resulting in intersecting or jagged paths due to the lack of geometry constraints. To overcome these limitations, we propose a novel neural path representation by designing a dual-branch Variational Autoencoder (VAE) that learns the path latent space from both sequence and image modalities. By optimizing the combination of neural paths, we can incorporate geometric constraints while preserving expressivity in generated SVGs. Furthermore, we introduce a two-stage path optimization method to improve the visual and topological quality of generated SVGs. In the first stage, a pre-trained text-to-image diffusion model guides the initial generation of complex vector graphics through the Variational Score Distillation (VSD) process. In the second stage, we refine the graphics using a layer-wise image vectorization strategy to achieve clearer elements and structure. We demonstrate the effectiveness of our method through extensive experiments and showcase various applications. The project page is \url{https://intchous.github.io/T2V-NPR}.

\end{abstract}


\begin{teaserfigure}
  \includegraphics[width=\textwidth]{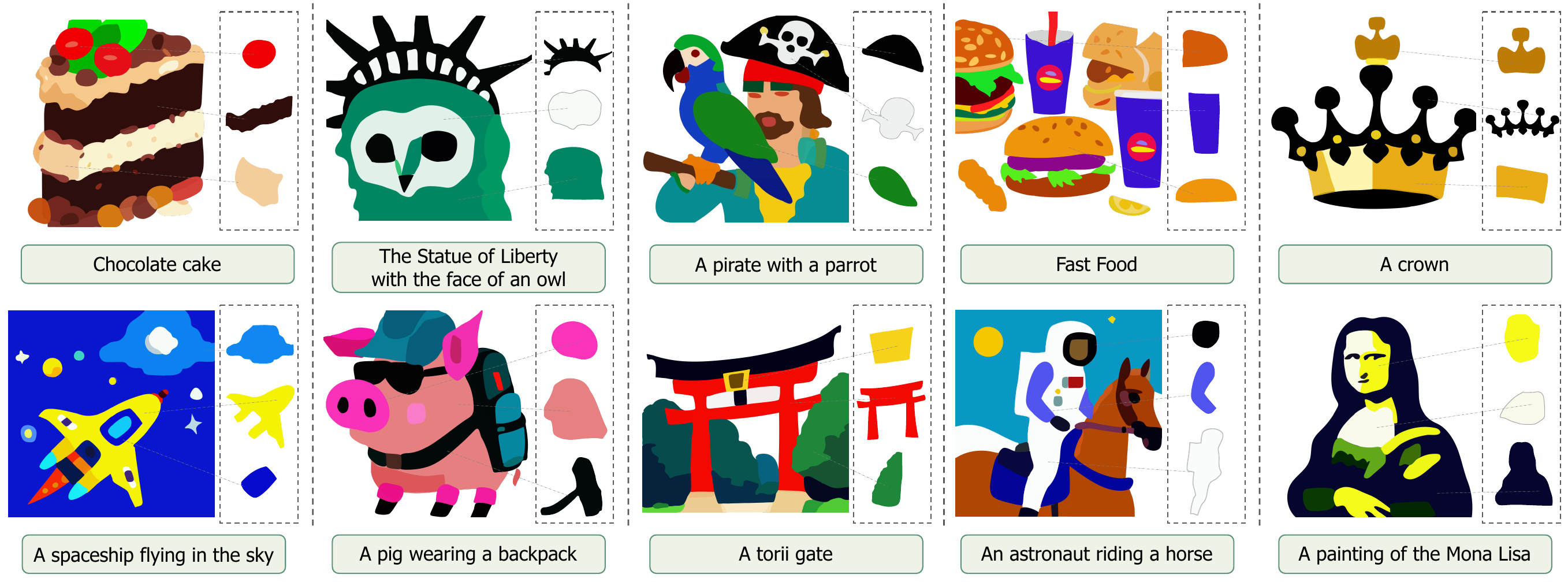}
  \caption{ Examples of text-guided vector graphics generated by our framework, with clear and valid layer-wise vector paths. }
  \label{fig:teaser}
\end{teaserfigure}

\maketitle

\section{Introduction}
\label{sec:introduction}

Vector graphics, specifically in the Scalable Vector Graphics (SVG) format, play an essential role in digital arts such as clipart, animation, and graphic design. Benefiting from their composition of geometric shapes, vector graphics are widely favored by designers due to the ease of manipulation, with the nature of resolution independence and compact file sizes. However, crafting high-quality vector graphics requires both professional expertise and considerable time investment. With the success of text-to-image (T2I) generation models \cite{rombach2022high, ruiz2022dreambooth}, recent works have started to explore text-to-vector graphics generation (T2V), aiming to make the creation more accessible to users with text prompts.

One common T2V approach is conducting existing image vectorization \cite{selinger2003potrace, dominici2020polyfit, ma2022towards} based on T2I results. However, T2I models often generate raster images with photographic and realistic styles, with intricate textures, and complex color variations. This poses a challenge when transitioning to vector graphics, where the goal is to achieve smooth geometric shapes and uniform colors. The conversion process often results in excessively complex vector elements, introducing complications for subsequent graphic manipulations and deviating from the simplicity and clarity expected in vector graphics. 
In recent developments, a new category of T2V methods has emerged (\eg, CLIPDraw \cite{frans2022clipdraw} and VectorFusion \cite{jain2022vectorfusion}). These methods directly optimize vector graphics paths using large pre-trained visual-textual models \cite{rombach2022high}. They represent vector graphics through parametric shape primitives (\eg, cubic Bézier curves) and refine path parameters, such as control points. However, directly optimizing control points often leads to intersecting or jagged paths, due to their high degrees of freedom and the lack of geometry constraints.

Therefore, in addition to representing vectors as the explicit parametric paths, it becomes imperative to develop an efficient representation to capture both geometric relationship and shape perception. Previous works have explored learning such latent representations based on sequences of SVGs using different network architectures, such as Recurrent Neural Networks (RNN) \cite{ha2017neural, lopes2019learned, wang2021deepvecfont} and Transformers \cite{carlier2020deepsvg, wang2023deepvecfont}. However, given the inherent complexity and diverse nature of SVGs, learning a comprehensive global SVG-level representation is challenging. As a result, previous methods are often restricted to generating SVGs within specific categories, such as~fonts \cite{lopes2019learned, wang2021deepvecfont, wang2023deepvecfont}, sketches \cite{ha2017neural, ribeiro2020sketchformer}, and simple icons \cite{carlier2020deepsvg, cao2023svgformer, wu2023iconshop}.

To overcome the above limitations, instead of learning global SVG features, we propose a novel neural path representation that effectively learns valid geometric properties of paths. This is motivated by the finding that complex vector graphics are often composed of simple paths \cite{chen2023editable, liu2023dualvector}. This compact representation ensures both simplicity and expressivity. To learn a neural path representation, we design a dual-branch Variational Autoencoder (VAE) to learn from both sequence and image modalities. They jointly optimize the shared path representation, and the sequence modality provides supervision for learning geometry properties, while the image modality helps to learn rendering visual features.

We further propose a two-stage path optimization method for conducting text-to-vector generation based on the learned path latent space. The underlying motivation is that obtaining high-quality vectors within one stage can be hard. In the first stage, we rely on the generation power of a large pre-trained diffusion model \cite{rombach2022high} for generating an initial SVG. Rather than using Score Distillation Sampling (SDS) loss \cite{poole2022dreamfusion} as in VectorFusion \cite{jain2022vectorfusion}, which may suffer from over-saturated, over-smooth, and low-diversity issues, we borrow the idea of Variational Score Distillation (VSD) \cite{wang2023prolificdreamer} to optimize a combination of neural paths given the text prompt. In the second stage, to further improve the geometry clarity and layer-wise structure of the initially generated SVG, we apply a path simplification and layer-wise optimization strategy to hierarchically enhance the paths.

We evaluate our approach through extensive experiments using vector-level, image-level, and text-level metrics. The results demonstrate the effectiveness of our model in generating high-quality and diverse vector graphics with valid paths and layer properties, given the input text prompt. \reffig{teaser} shows examples of text-to-vector generation results produced by our framework. Our model enables various applications except for T2V generation, such as vector graphics customization, image-to-SVG generation, and SVG animation. Our key contributions are:
\begin{itemize}[leftmargin=*]
  \item We introduce a novel T2V generation pipeline, innovated by the idea of optimizing local neural path representation for high-quality vector graphics generation.
  \item We propose a dual-branch VAE for learning neural path representation from both sequence and image modalities. 
  \item We develop a two-stage text-driven neural path optimization method to guide the creation of vector graphics with valid and layer-wise SVG paths.
  \item We demonstrate several practical applications enabled by our pipeline.
\end{itemize}

\begin{figure*}[tbp]
  \centering
  \includegraphics[width=1.0\linewidth]{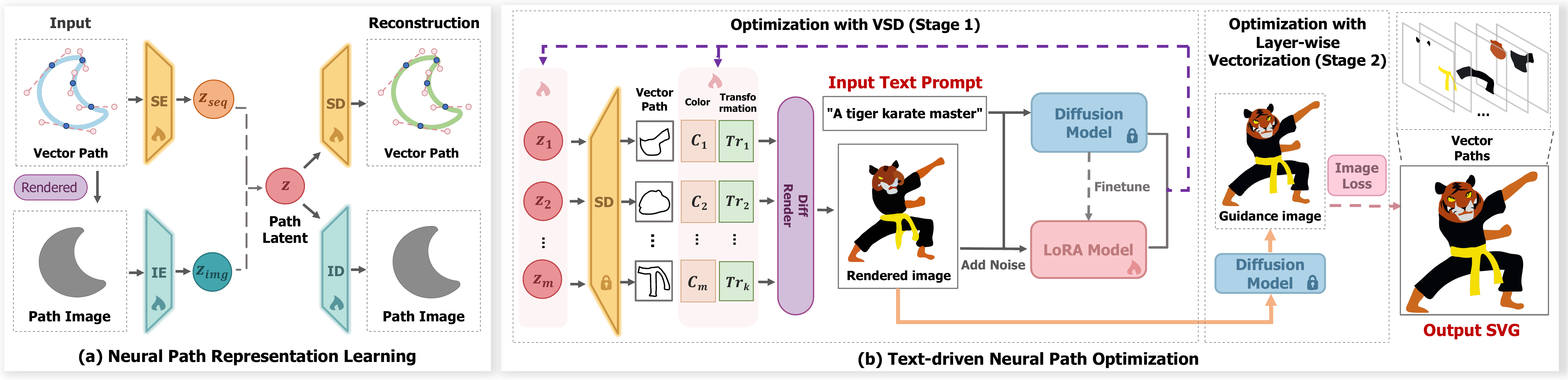}
  \caption{ \label{fig:pipeline} Our system pipeline starts with learning a neural representation of paths by training a dual-branch VAE. Next, we optimize the SVG, represented with neural paths, to align with the provided text prompt, which is achieved through a two-stage path optimization process. }
\end{figure*}

\section{Related Work}
\label{sec:related_work}

\subsection{Diffusion Models for T2I Generation}  
\label{sec:diffusion_models}
Recently, diffusion models (DM) have become state-of-the-art in T2I generation. It is a family of generative models, involving a forward process of perturbing the data with noise and a reverse process that gradually adds structure from noise \cite{sohl2015deep, ho2020denoising, nichol2021glide}. 
Stable Diffusion \cite{rombach2022high} further introduces image latent to the diffusion model, overcoming resolution limitations and enabling the generation of amazing high-resolution results.
The diverse type of user-defined condition modalities enable numerous downstream applications, such as sketch-guided image generation \cite{zhang2023adding, voynov2023sketch}, multi-modal conditions in image synthesis \cite{zhan2023multimodal}, and various image-to-image generation tasks \cite{saharia2022palette, zhang2023inversion}.
Diffusion models also support flexible text-guided image editing \cite{meng2021sdedit, hertz2023delta} and customization tasks \cite{ruiz2022dreambooth, kumari2022multi}.
To generate personalized images, the pioneering work DreamBooth \cite{ruiz2022dreambooth} refines the weights of a diffusion model with a unique token to capture distinctive characteristics within a set of images. Subsequent works focus on fine-tuning only particular parts of the network, including low-rank weight residuals \cite{hu2021lora} and cross-attention layers \cite{kumari2022multi}.
Rather than aiming at raster image generation, our method leverages the powerful pre-trained T2I model as priors to guide the generation of vector graphics.

\subsection{Text-to-Vector Generation} 
\label{sec:text_to_vector_generation}
One approach in T2V generation combines T2I generation with image vectorization methods. 
Traditional vectorization methods involve segmenting images into regions based on color similarity \cite{kopf2011depixelizing} and fitting curves to the region boundaries \cite{selinger2003potrace, yang2015effective, favreau2017photo2clipart, hoshyari2018perception,dominici2020polyfit}.
The advent of differentiable rendering techniques \cite{li2020differentiable} enhances image vectorization by enabling the use of loss functions in image space. LIVE \cite{ma2022towards} optimizes path parameters guided by the reconstruction of the input image. Recent developments have explored training end-to-end neural networks for image vectorization \cite{reddy2021im2vec, chen2023editable, rodriguez2023starvector}.
However, this kind of method relies on images generated by pre-trained T2I models, which struggle to produce high-quality SVG-style images featuring clear geometric primitives and flat colors.
With the development of generative AI, several community-made tools and commercial products such as Adobe Illustrator \cite{adobe-illustrator}, Illustroke \cite{illustroke}, and Kittl \cite{kittl}, offer capabilities to generate vector graphics from text prompts by employing vectorization techniques with T2I methods. To produce visually appealing SVG-style images, it is essential to fine-tune pre-trained T2I models \cite{hu2021lora,ruiz2022dreambooth}. These fine-tuning methods heavily rely on the training dataset, requiring a labor-intensive collection of images and textual tags.

Another approach for T2V generation involves directly optimizing the geometric and color parameters of SVG paths under the guidance of pre-trained vision-language models, such as the CLIP model \cite{radford2021learning} or the diffusion model \cite{rombach2022high}. CLIP-based methods \cite{frans2022clipdraw, schaldenbrand2022styleclipdraw, song2022clipvg} utilize the image-text similarity metric within CLIP latent space to create vector graphics from text prompts. 
CLIPasso \cite{vinker2022clipasso} converts images to object sketches using a CLIP perceptual loss, preserving the semantics and structure of the subject.
Apart from the CLIP distance, Score Distillation Sampling (SDS) loss based on a T2I diffusion model is used for optimizing SVG to align with text prompts across various applications such as fonts \cite{iluz2023word}, vector graphics \cite{jain2022vectorfusion, xing2023svgdreamer}, and sketches \cite{xing2023diffsketcher, gal2023breathing}.
However, these methods often lead to SVGs with cluttered and irregular paths, as the direct optimization of control points in parametric paths like cubic Bézier curves lacks essential geometric relationships.

\subsection{Neural Representation for SVG} 
\label{sec:neural_representation_for_svg}
Previous works have explored learning various representations of SVG, and designing different network architectures to understand the geometric information and global perception inherent in SVG data.
SketchRNN \cite{ha2017neural} leverages a recurrent neural network (RNN) to generate vector paths for sketches. 
SVG-VAE \cite{lopes2019learned} uses an image autoencoder to capture font style features and employs LSTMs to generate vector fonts.
Sketchformer \cite{ribeiro2020sketchformer} uses a Transformer to recover sketch strokes in a vector form based on raster images.
To maintain hierarchical relationships in SVGs, DeepSVG \cite{carlier2020deepsvg} employs a hierarchical Transformer-based network to generate vector icons composed of multiple paths.
Dual-modality learning framework \cite{wang2021deepvecfont, wang2023deepvecfont, liu2023dualvector} utilizes both sequence and image features to synthesize accurate vector fonts.
A Transformer-based framework is designed to vectorize line drawings with dual-modal supervision \cite{liu2022end}, but it lacks smooth interpolation characteristics in the latent space.
While these techniques have not yet supported text-guided SVG creation, IconShop \cite{wu2023iconshop} achieves T2V generation by representing SVGs as token sequences combined with text tokens.

Previous works focus on learning SVG-level latent representations from SVG command sequences. 
However, the vast diversity of path combinations poses a challenge in learning a universal global SVG-level representation.
As a result, existing methods predominantly generate SVGs in specific categories like fonts \cite{lopes2019learned, wang2021deepvecfont, wang2023deepvecfont}, sketches \cite{ha2017neural, ribeiro2020sketchformer}, and simple icons \cite{carlier2020deepsvg, cao2023svgformer, wu2023iconshop}. 
In contrast, our neural path representation captures valid path properties, while enabling the generation of diverse and complex vector graphics from text prompts.

\section{Overview}
\label{sec:overview}

Given a text prompt, our goal is to generate an SVG that aligns with the semantics of the text prompt while exhibiting desirable path properties and layer-wise structures consistent with human perception. Since an SVG is composed of a set of paths, denoted as $SVG = \{Path_1, Path_2,...,Path_m\}$, our objective is to optimize a set of $m$ paths based on a text prompt $T$ through: 

\paragraph{Neural Path Representation Learning (\refsec{neural_path_representation_learning})}
The path geometry consists of connected cubic Bézier curves. Our objective is to learn a neural path representation by mapping each path into a latent code denoted as $z$, which captures valid geometric properties. To achieve this, we propose a dual-branch VAE that learns from both the image and sequence modalities of paths (refer to \reffig{pipeline} (a)).

\paragraph{Text-driven Neural Path Optimization (\refsec{text_driven_neural_path_optimization})}
With the learned neural path representation, an SVG can be represented by a set of latent codes, along with color and transformation parameters associated with each path, denoted as $SVG = \{\theta_1, \theta_2,...,\theta_m\}$, where $\theta_i=(z_i,C_i,Tr_i)$. Here, $z_i$, $C_i$, and $Tr_i$ represent the latent code, color parameter, and transformation parameter for the $i$-path ($Path_i$), respectively. Next, we use the text prompt $T$ to optimize the combination of $m$ paths through two stages. In the first stage, we utilize a pre-trained diffusion model as a prior to optimize the combination of neural paths aligned with $T$. In the second stage, we apply a layer-wise vectorization strategy to optimize path hierarchy, ensuring clear visual elements and layer structures in the generated SVG. Finally, after decoding the optimized latent codes into Bézier curves, we obtain the resulting SVG (refer to \reffig{pipeline} (b)).

\section{Neural Path Representation Learning} 
\label{sec:neural_path_representation_learning}

Previous T2V approaches directly optimize the control points of parametric paths. However, this often leads to intersecting or jagged paths due to the high degrees of freedom and the lack of geometric constraints. To address this issue, we propose a novel approach that learns a neural path representation within a latent space, capturing the valid geometric properties of paths and enabling the optimization of paths while ensuring geometric regularity. For this purpose, we design and train a dual-branch VAE to learn a latent space for path representation.

\subsection{Dual-branch VAE}
\label{sec:dual_branch_vae}
A parametric path (ignoring the color parameter) can be defined as a series of cubic Bézier curves connected end-to-end, denoted as $Path=(p_1,p_2,\ldots,p_k)$, where $p_j$ for $j=1$ to $k$ represents the $k$ control points used to define the cubic Bézier curves. By utilizing the differentiable rasterizer $\mathcal{R}$ \cite{li2020differentiable}, we can render the vector path and obtain the rasterized path image $I= \mathcal{R}(Path)$.
The sequence representation of a path contains rich geometric information, such as the ordering and connections between points in the path. Training a VAE based on the point sequence enables effective learning of geometric properties, but it struggles to precisely reconstruct the rendered shape. On the other hand, the image representation is better at capturing visual features after rendering, but it cannot represent the relationships between control points, as different sequences of control points may result in the same rendered shape.
To address these challenges, we propose a dual-branch encoder-decoder VAE that learns a shared latent space from path data in both vector and image modalities, as shown in \reffig{pipeline} (a). This approach allows for the incorporation of both geometric and visual information, enabling a more comprehensive and accurate learning of paths.

\paragraph{Encoders}
The encoder of our dual-branch VAE is composed of a sequence encoder and an image encoder. The \textbf{sequence encoder} takes the control point sequence as input and employs a transformer architecture with an attention mechanism to exploit the geometric relationships between control points. First, we normalize the control points sequence to the range of $[0,1]$ and then use a learnable embedding layer to project each normalized control point into a $d_h$-dimensional vector. Similar to DeepSVG \cite{carlier2020deepsvg}, we use positional encoding to embed the index of each point in the sequence. The sequence embedding $e_{seq} \in \mathbb{R}^{d_h \times k}$ is then fed into the six transformer layers. Each layer consists of masked multi-head attention and feed-forward layers \cite{vaswani2017attention}. Finally, a linear projection is applied to obtain the aggregated output sequence features $z_{seq} \in \mathbb{R}^{d_S}$. To further learn the shape perception of the path, we also adopt an \textbf{image encode}r composed of six convolutional layers. It takes the rasterized path image $I$ as input and outputs the feature $z_{img} \in \mathbb{R}^{d_I}$. In our implementation, we set $d_h=64$, $d_S=32$, and $d_I=64$.

\paragraph{Modality Fusion}
We fuse the sequence and image features to create a comprehensive representation of paths. Specifically, the sequence features $z_{seq}$ and image features $z_{img}$ are concatenated and passed through a linear projection layer, obtaining a latent code $z \in \mathbb{R}^{d_F}$, where $d_F=24$, that is shared with both modalities.

\paragraph{Decoder}
The latent code $z$ is passed through the two decoding branches to reconstruct the vector path and path image, respectively. The \textbf{sequence decoder} has a similar Transformer-based architecture to the sequence encoder. It takes $z$ as input and outputs the decoded points sequence $\hat{Path} = (\hat{p_{1}}, \hat{p_{2}}, \ldots, \hat{p_{k}})$. The \textbf{image decoder} is a deconvolutional neural network that utilizes $z$ to generate the reconstructed path image $\hat{I}$.

\begin{figure}[tbp]
  \centering
  \includegraphics[width=0.9\columnwidth]{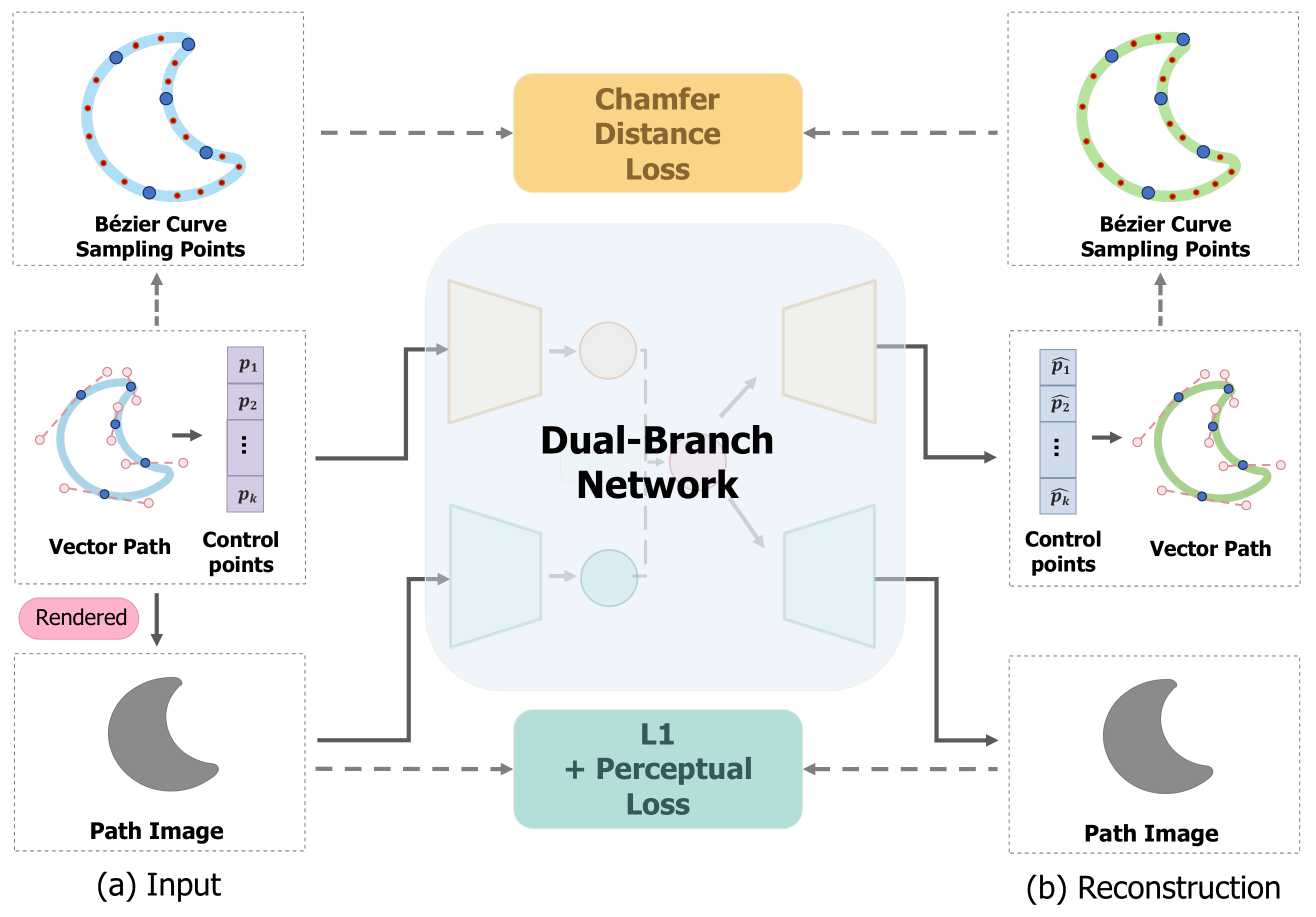}
  \caption{ \label{fig:train_loss} Training loss of the dual-branch VAE. }
\end{figure}

\subsection{Training}
\label{sec:training_details}

\paragraph{Loss Function}
The dual-branch VAE is trained end-to-end to reconstruct the input using dual-modality losses. However, simply using mean squared error (MSE) loss between the input and output control point sequences for sequence reconstruction can lead to overfitting, as different control point sequences can produce similar shapes. To address this, we introduce a shape-level loss to capture shape features. As depicted in \reffig{train_loss}, we first sample $n$ auxiliary points (in our implementation, $n=4$) from each cubic Bézier curve of a vector path, resulting in the auxiliary points set $P_{aux}$. Next, we calculate the Chamfer Distance between the auxiliary points set of the input path $Path$ and that of the reconstructed path $\hat{Path}$ by summing the distances between the nearest correspondences of the two point sets. This calculation yields the Chamfer loss $\mathcal{L}_{cfr}(Path,\hat{Path})$. 
The image-level loss combines the $L_1$ loss and perceptual loss between the input image $I$ and reconstructed images $\hat{I}$. Specifically, it can be expressed as $\mathcal{L}_{img} = |I - \hat{I}|_{1} + \mathcal{L}_{percep}(I, \hat{I})$.
In addition to the reconstruction losses in both modalities, the latent space is also regularized by the KL divergence loss $\mathcal{L}_{kl}$, which encourages the latent code $z$ to follow a Gaussian distribution $\mathcal{N}(0, I)$.

The overall loss function is:
\begin{equation}
\mathcal{L}=\lambda_{cfr}\mathcal{L}_{cfr}+\lambda_{img}\mathcal{L}_{img}+\lambda_{kl}\mathcal{L}_{kl},
\end{equation}
where $\lambda_{cfr}$, $\lambda_{img}$, $\lambda_{kl}$ are the weights of each loss term. We set $\lambda_{cfr}=1$, $\lambda_{img}=0.1$, $\lambda_{kl}=0.01$ in our implementation.

\begin{figure}[tbp]
  \centering
  \includegraphics[width=1.0\columnwidth]{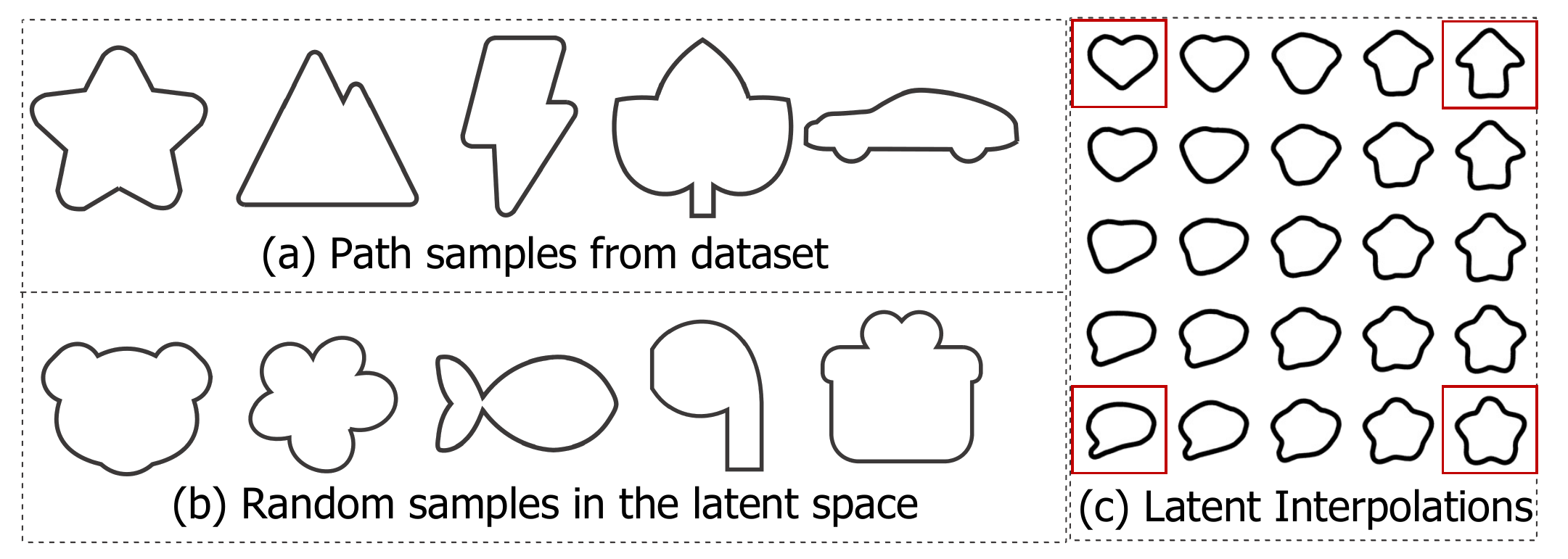}
  \caption{ \label{fig:path_latent} Path samples and latent interpolations. (a) Path examples from the dataset. (b) Path samples decoded from random vectors in the latent space. (c) Latent interpolations among four given samples (marked in red). }
\end{figure}

\paragraph{Training Details}
We train our dual-branch VAE on the FIGR-8-SVG dataset \cite{clouatre2019figr}, which consists of black-and-white vector icons. To preprocess the data, we follow the same steps as IconShop \cite{wu2023iconshop} to obtain valid SVG data. We extract paths from the SVGs and remove duplicate shapes. In \reffig{path_latent}, we showcase several examples of paths from the dataset.
For the raw input, since the control point sequences in each path can have variable lengths, we pad the point sequences with zeros to a fixed length (in our implementation, $k=50$) and filter out those with longer lengths. This results in 200,000 samples for model training. The image resolution for training VAE is \(64\times64\).
To train the model, we use the Adam optimizer with an initial learning rate of 0.001. We incorporate linear warmup and decay techniques. The dropout rate in all transformer layers is set to 0.1. We train the dual-branch network for 100 epochs.

Upon completion of training, we obtain a smooth latent space shared by both modalities. In \reffig{path_latent}, we present paths decoded from random samples in the learned latent space, along with smooth path interpolations.

\section{Text-driven Neural Path Optimization}
\label{sec:text_driven_neural_path_optimization}

With the learned neural path representation, an SVG can be represented by a combination of $m$ paths in the latent space, denoted as $SVG = \{\theta_1, \theta_2,...,\theta_m\}$, where $\theta_i=(z_i,C_i,Tr_i)$. Here, the latent code $z_i$ defines the shape, $C_i$ defines the color, and $Tr_i$ defines the affine transformation of the $path_i$. We initialize the latent codes by randomly drawing from a zero-mean Gaussian distribution. Our aim is to optimize these parameters based on the given text prompt $T$. Unlike previous methods \cite{jain2022vectorfusion, xing2023svgdreamer} that optimize control points explicitly, we optimize the latent code $z$ within the learned space to guarantee the regularity of each path in the result. After optimization, we can obtain control points of a path by decoding $z_i$ with our sequence decoder and transforming points with $Tr_i$, as $Path_i={Tr_i\cdot SeqDec(z_i)}$.

Moreover, previous works \cite{jain2022vectorfusion, frans2022clipdraw} simultaneously optimize all paths of an SVG in a single stage, resulting in cluttered paths and a messy layer structure. To tackle this challenge, we have developed a two-stage neural path optimization process, as shown in \reffig{pipeline} (b). In the first stage, we employ VSD defined on a pre-trained text-to-image diffusion model to optimize paths, resulting in an initial SVG that aligns with $T$. In the second stage, starting from the initial SVG, we apply a layer-wise vectorization strategy to hierarchically optimize path combinations, ensuring clear visual elements and well-defined layer-wise structures in the result SVG. Next, we introduce these two stages.

\subsection{Optimization with Variational Score Distillation}
\label{sec:optimization_with_variational_score_distillation}
In this stage, we optimize an initial SVG guided by the text prompt $T$. We draw inspiration from VectorFusion \cite{jain2022vectorfusion}, which leverages a pre-trained text-to-image diffusion model as a prior to optimize path parameters through a Score Distillation Sampling (SDS) process. The process begins with an $SVG$ with randomly initialized path parameters $\theta$, and in each iteration, the $SVG$ is rendered using the differentiable rasterizer $\mathcal{R}$ to obtain a raster image $I_{SVG}=\mathcal{R}(SVG)$. The pre-trained encoder of the diffusion model encodes $I_{SVG}$ into latent features $\mathbf{x}=\mathcal{E}(I_{SVG})$, and a Gaussian noise $\epsilon \in \mathcal{N}(0, I)$ is added to $\mathbf{x}$, obtaining $\mathbf{x}_t$ at time $t$ of the forward diffusion process. Finally, the SDS loss is defined as the distance between the added noise $\epsilon$ and the predicted noise $\epsilon_{\phi}$ using the pre-trained diffusion model, and its gradient to optimize $\theta$ can be estimated as follows:
\begin{equation}
    \nabla_{\theta} \mathcal{L}_{\mathrm{SDS}} \triangleq \mathbb{E}_{t, \epsilon}\left[w(t)\left(\epsilon_{\phi}\left(\mathbf{x}_{t} ; T, t\right)-\epsilon\right) \frac{\partial \mathbf{x}_{}}{\partial \theta}\right]
\end{equation}
where $w(t)$ is a time-dependent weighting function.

Despite its success, empirical observations have revealed that results optimized from SDS suffer from issues such as over-saturation, over-smoothing, and a lack of diversity. These issues stem from SDS treating parameter $\theta$ as a single point and using a single point to approximate a distribution output by the diffusion model. In light of this, we leverage the VSD loss proposed by ProlificDreamer \cite{wang2023prolificdreamer} to replace SDS in our optimization. Unlike SDS, VSD models the parameter $\theta$ as a distribution, and consequently, the images rendered by $SVG$ with parameter $\theta$ are also a distribution. Following ProlificDreamer, we employ LoRA (Low-rank adaptation) \cite{hu2021lora} of the pre-trained diffusion model to model the distribution of the rendered images. Therefore, the VSD loss is defined as the distance between the noises predicted by the pre-trained diffusion model and the LoRA model. Its gradient to optimize $\theta$ can be formulated as follows:

\begin{equation}
    \nabla_\theta \mathcal{L}_{\mathrm{VSD}} \triangleq \\
    \mathbb{E}_{t, \epsilon}\left[w(t)\left(\epsilon_\phi\left(\mathbf{x}_t ; T, t\right)-\epsilon_{\mathrm{LoRA}}\left(\mathbf{x}_t ; T, t\right)\right) \frac{\partial \mathbf{x}}{\partial \theta}\right]
\end{equation}

The use of VSD helps generate SVGs with higher quality and diversity. We denote the SVG optimized from this stage as $SVG^0$, which will guide the layer-wise refinement in the next stage.

\begin{figure}[tbp]
  \centering
  \includegraphics[width=1.0\columnwidth]{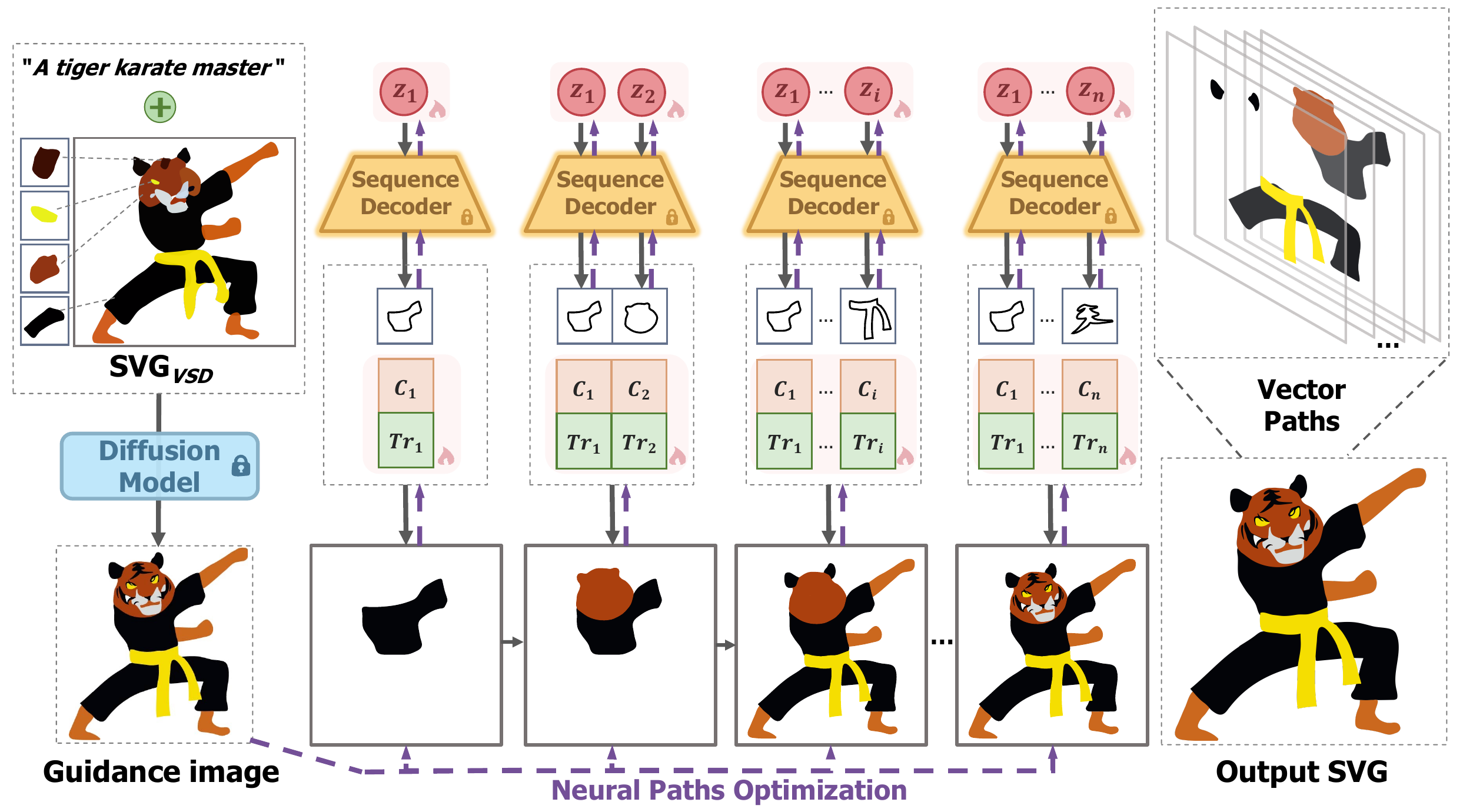}
  \caption{ \label{fig:layerwise} Layer-wise optimization strategy. }
\end{figure}

\subsection{Optimization with Layer-wise Vectorization}
\label{sec:optimization_layer_wise_vectorization}

Although VSD optimization is effective in aligning SVGs with text prompts, it often results in cluttered and stacked paths. This can introduce artifacts and lead to a poorly organized layer structure within the SVG, complicating subsequent editing and modification. To enhance the clarity of vector elements and the hierarchical structure of the generated SVG, we introduce a second-stage optimization based on $SVG^0$ obtained from the previous stage. This stage incorporates a path simplification step to obtain a simplified path set, and a layer-wise vectorization strategy to hierarchically optimize path combinations with the assistance of a guidance image. This stage can improve the overall quality and organization of the resulting SVG, making it easier to edit and reuse.

\paragraph{Path Simplification}
We remove paths in $SVG^0$ with an opacity below 0.05 or an area smaller than 10 pixels, and merge overlapping paths that exhibit the same color into a single path using an overlap threshold of 5 pixels.
These steps reduce the number of paths from $m$ to $n$. The simplified path set with sequence representation is then inversely transformed into the latent space using our sequence encoder, obtaining a new set of latent codes and resulting in a new SVG, $\tilde{SVG^0}$.

\paragraph{Layer-wise Optimization Strategy}
After simplification, we further refine the $\tilde{SVG^0}$ with enhanced perceptual clarity and a better layer-wise structure with the assistance of a guidance image. The guidance image is obtained by first rendering $\tilde{SVG^0}$ into the image $I_{SVG}^0$. Then, we perturb $I_{SVG}^0$ with Gaussian noise, setting the noise strength to 0.4, and gradually remove the noise using a pre-trained diffusion model. This process helps reduce artifacts in the initial SVG and yields a guidance image $I_g$ with clearer and more precise visual elements. 

With the help of $I_g$, we introduce a layer-wise optimization strategy. Specifically, we sort the paths in $\tilde{SVG^0}$ by area and then optimize the top $k$ paths with the largest areas in each iteration. A recursive pipeline progressively adds paths according to a path number schedule, thus optimizing the SVG from coarse to fine, as shown in \reffig{layerwise}.

In each iteration, we optimize the parameters of the top $k$ paths in $\tilde{\theta^0}$, denoted as $SVG^k = \{\theta_1, \theta_2, ..., \theta_k\}$. The optimization is based on CLIP loss and IoU loss defined between ${SVG}^k$ and the guidance image $I_g$. The CLIP loss is computed by summing the $L_2$ distances between the intermediate-level activations of the CLIP model as follows:
\begin{equation}
  L_{CLIP} = \sum_{l} \lVert CLIP_l(\mathcal{I}_g) - CLIP_l(\mathcal{R}(SVG^k)) \rVert _2^2,
\end{equation}
$CLIP_l$ denotes the CLIP encoder activation at layer $l$. It evaluates the image-level similarity and encourages the rendering of $SVG^k$ to be faithful to the guidance image $I_g$. Moreover, to encourage a limited number of paths to cover the content of the guidance image as much as possible, we apply the intersection-over-union (IoU) loss between the rendered silhouette of ${SVG}^k$ and the foreground region of $I_g$. The overall loss function is defined as $L_{lyr} = L_{CLIP} + \lambda_{IoU} L_{IoU}$, where $\lambda_{IoU}$ is set to 0.01.

As the iterations progress, more paths are involved in the optimization process, and the SVG gradually adds details. When all paths are optimized, we obtain the final result, $SVG^n$ with $n$ paths.

\section{Experiments}
\label{sec:experiment}

\paragraph{Experiment Setup}
To evaluate our method, we collect 160 text prompts from the Stable Diffusion Prompts dataset \cite{dehouche2023text}, including a diverse range of characters, actions, and scenes. For each prompt, we generate 5 SVGs, culminating in a total of 800 vector graphics. To encourage the generation towards a flat vector style, we append the phrase "minimal flat 2d vector" to each prompt. In VSD optimization process, we utilize the official "SD-v1-5" checkpoint\footnote{https://huggingface.co/runwayml/stable-diffusion-v1-5} with a guidance scale of 10, and timestep \(t\sim\mathcal{U}(50,950)\) is uniformly sampled.

\subsection{Evaluation Metrics}
\label{sec:evaluation_metrics}
We evaluate the quality of our results from vector-level, image-level, and text-level perspectives.

\paragraph{Vector-level}
Drawing upon criteria from prior perception research in vector graphics \cite{favreau2017photo2clipart, dominici2020polyfit}, a good SVG should minimize redundant paths to preserve compactness and editability. Based on this guideline, we assess vectorization quality using the following metrics:
(a) \textbf{Smoothness}: Inverse of the average curvature variation of the paths in generated SVGs.
(b) \textbf{Simplicity}: The average number of paths in generated vector graphics.
(c) \textbf{Layer-wise semantics}: The semantics of paths are evaluated by comparing the decrease in CLIP similarity between the SVG render results and corresponding text prompts before and after randomly removing 30\% of the SVG paths. A larger decrease indicates that each path has more specific semantics.

\paragraph{Image-level}
To evaluate the visual quality and diversity of vector graphics, we collected a dataset of 800 well-designed vector graphics from \textit{iconfont}\footnote{https://iconfont.cn}, encompassing various categories including characters, animals, and scenes. These samples serve as the ground truth for calculating the FID metrics on rendered images of SVGs.

\paragraph{Text-level}
To compute whether the generated SVG is aligned with the input text prompt, we define the text-level similarity by calculating the CLIP cosine similarity \cite{radford2021learning} between the text prompt and rendered SVG.

\subsection{Baselines}
\label{sec:comparison_with_existing}

We compare our proposed pipeline with two types of T2V pipelines. 
\paragraph{Vectorization with T2I} 
We first generate images from text prompts using a diffusion model and then convert the images into SVGs using two distinct vectorization methods:
(a) \textbf{Potrace} \cite{selinger2003potrace}: A traditional vectorization method involves image segmentation and curve fitting to transform raster images into SVGs.
(b) \textbf{LIVE} \cite{ma2022towards}: A deep learning method that generates SVGs by optimizing paths using loss functions defined in the image space. We use the same number of paths as our method.

\paragraph{Text-guided SVG optimization}
We compare our method with CLIP-based and diffusion-based optimization approaches:
(c) \textbf{CLIPDraw} \cite{frans2022clipdraw}: A method that leverages the image-text similarity metric of CLIP to optimize SVGs from prompts.
(d) \textbf{Diffsketcher} \cite{xing2023diffsketcher}: A method incorporates both image-level CLIP loss and SDS loss for text-guided sketch generation.
(e) \textbf{VectorFusion} \cite{jain2022vectorfusion}: This approach employs SDS loss to optimize SVGs to align with given text prompts. It offers two optimization routes: starting from scratch or refining vectorized results produced by LIVE.
We use 128 paths for text-guided SVG optimization methods, consistent with the default setting in VectorFusion.

\begin{table}[tbp]
  \caption{ Quantitative comparison with existing methods. }
  \small
  \resizebox{\linewidth}{!}{
    \begin{tabular}{|c|c|c|c|c|c|}
      \hline
      Methods                     & Smooth $\uparrow$ & Simp $\downarrow$ & Layer $\uparrow$ & FID $\downarrow$ & CLIP $\uparrow$ \\ \hline
      T2I + Potrace               & 0.7846            & 430               & 0.0226           & 104.92           & 0.2732          \\ \hline
      T2I + LIVE                  & 0.5797            & 40                & 0.0119           & 97.82            & 0.2519          \\ \hline
      T2I + LoRA + Potrace        & 0.7882            & 160               & 0.0395           & \textbf{45.16}   & 0.2729          \\ \hline
      CLIPDraw                    & 0.4711            & 128               & 0.0317           & 123.55           & 0.2832          \\ \hline
      Diffsketcher                & 0.4829            & 128               & 0.0105           & 92.36            & 0.2607          \\ \hline
      Vectorfusion (from scratch) & 0.6322            & 128               & 0.0139           & 65.71            & 0.2675          \\ \hline
      Vectorfusion (from LIVE)    & 0.5025            & 128               & 0.0258           & 76.45            & 0.2880          \\ \hline
      \textbf{Ours}               & \textbf{0.8012}   & \textbf{40}       & \textbf{0.0591}  & 52.30            & \textbf{0.3015} \\ \hline
    \end{tabular}
  }
  \label{tab:table_quality_eval}
\end{table}

\subsection{Comparisons}
\begin{figure}[tbp]
  \centering
  \includegraphics[width=1.0\columnwidth]{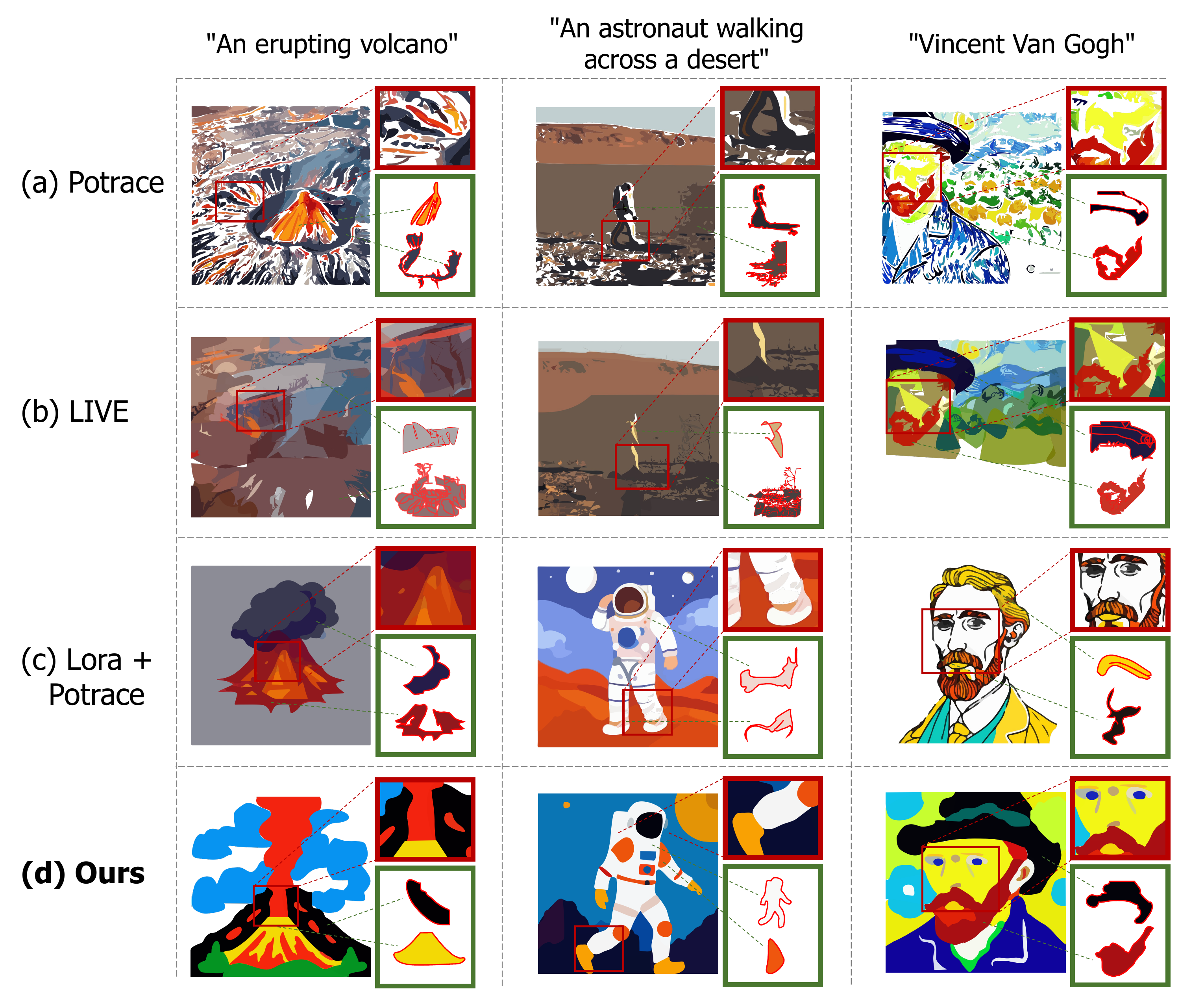}
  \caption{ \label{fig:result_t2i2v} Qualitative comparison to vectorization with T2I methods. }
\end{figure}

\begin{figure*}[tbp]
  \centering
  \includegraphics[width=1.0\linewidth]{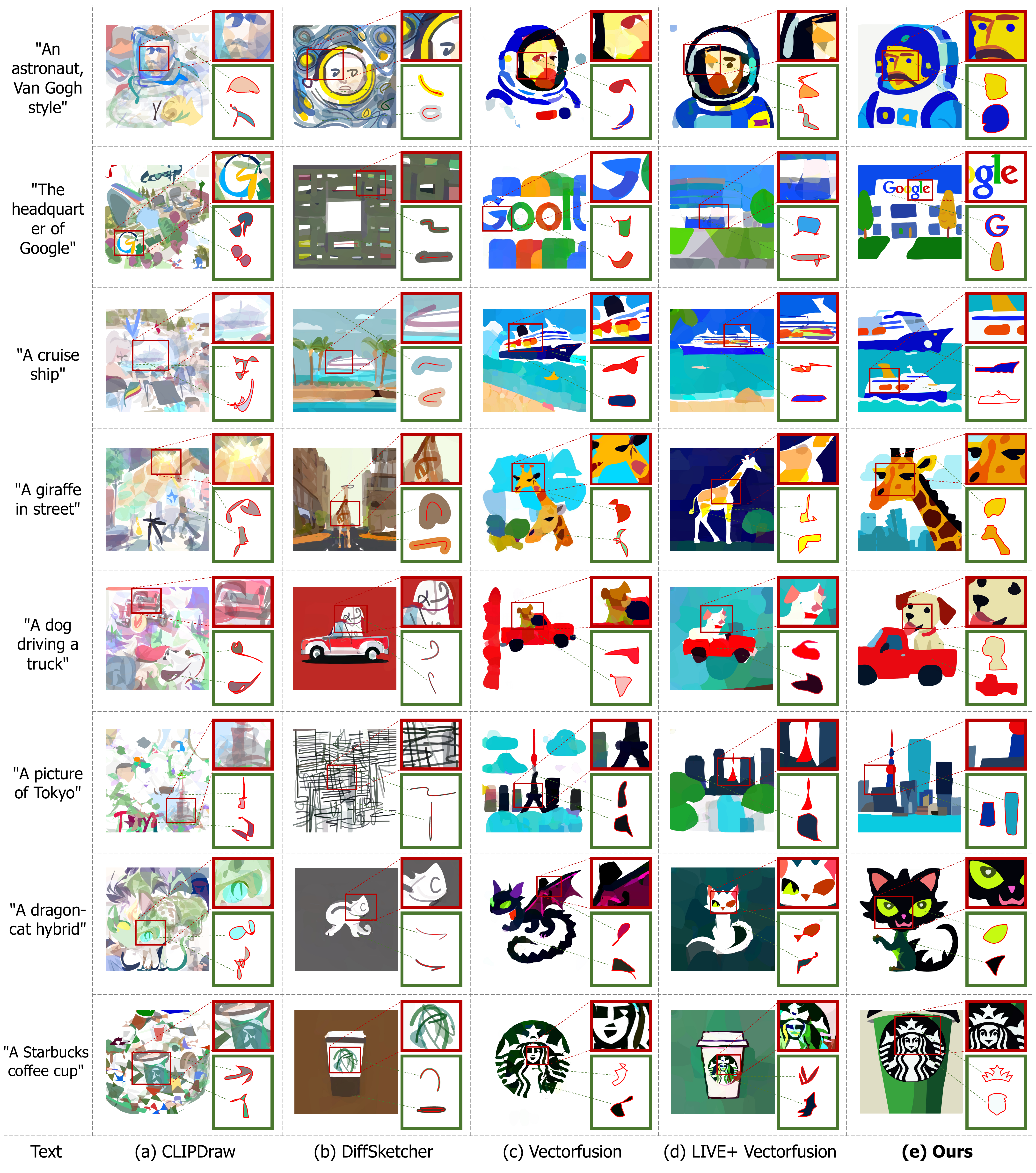}
  \caption{ \label{fig:result_t2v} Qualitative comparison with text-guided SVG optimization methods. }
\end{figure*}

We evaluate the performance of our method by comparing it with baselines qualitatively and quantitatively. The quantitative results are provided in \reftab{table_quality_eval} and the qualitative results are shown in \reffig{result_t2i2v} and \reffig{result_t2v}. As shown in \reftab{table_quality_eval}, our method outperforms other approaches from a comprehensive perspective.

\paragraph{Comparisons with Vectorization with T2I Methods}
Though with a vector style prompt suffix, T2I models still often generate raster images with intricate textures and complex color variations. It poses a challenge when converting to vector graphics with smooth geometric shapes and uniform colors.
As shown in \reffig{result_t2i2v}, the vectorization results using Potrace exhibit overly complex vector elements.
This leads to a high number of paths (indicated by a high simplicity score in \reftab{table_quality_eval}) and a lack of layer organization among these paths, resulting in diminished path semantics (reflected by a low layer-wise semantics score in \reftab{table_quality_eval}). Furthermore, such vectorization results often deviate from the style of well-designed SVGs, as evidenced by the low FID scores in \reftab{table_quality_eval}.
LIVE faces similar challenges. When the path number is set the same as our method, LIVE obtains suboptimal vectorization outcomes as the constrained path number is often insufficient for accurately reconstructing the images generated by the T2I model. Furthermore, the SVGs produced by LIVE exhibit numerous irregular and broken paths, as indicated by the low smoothness score in \reftab{table_quality_eval}.

LoRA \cite{hu2021lora} is a popular technique to fine-tune a pre-trained T2I model for specific styles. To produce SVG-style images, we utilize the "Vector Illustration" LoRA\footnote{https://civitai.com/models/60132/vector-illustration} fine-tuned on the base diffusion model "SD-v1-5", and then convert the generated images to vector graphics using Potrace. Since this LoRA model is specifically fine-tuned on well-designed vector graphics images similar to the iconfont dataset we used for evaluation, the generated results exhibit lower FID scores. However, as demonstrated in \reffig{result_t2i2v} (c), while this baseline achieves pleasing visual results, it still struggles with overcomplicated and disorganized paths, as this is a common issue in traditional image vectorization methods. In contrast, our methods produce visually appealing SVGs with smooth paths and layer-wise structures.

\paragraph{Comparisons with Text-guided SVG Optimization Methods}
These methods directly optimize the control points of parametric paths. However, due to their high degrees of freedom and the lack of geometry constraints, the control points often undergo complex transformations to generate SVGs that align with the text prompts. This results in low path smoothness, as indicated by the lower Smooth score in \reftab{table_quality_eval}. Zoom-in illustrations in \reffig{result_t2v} highlight the issues of intersecting and jagged paths, leading to visually unappealing outcomes. Moreover, the resulting SVGs often contain complex and redundant shapes, making them difficult to edit. In contrast, our neural path representation effectively captures valid geometric properties of paths. By enabling text-guided optimization within a constrained latent space, our approach facilitates the generation of SVGs with smooth paths. 

Additionally, VectorFusion, which simultaneously optimizes all paths in a single stage, often leads to cluttered paths and a disorganized layer structure. In contrast, our pipeline adopts a two-stage text-driven neural path optimization method, resulting in vector graphics with clear and valid layer-wise vector paths. Despite using fewer paths, our method still produces SVGs that maintain semantic alignment with the text prompts. This indicates that our paths have better semantic meaning, which is further supported by the higher layer-wise semantics score in \reftab{table_quality_eval}.

\begin{figure}[tbp]
  \centering
  \includegraphics[width=\columnwidth]{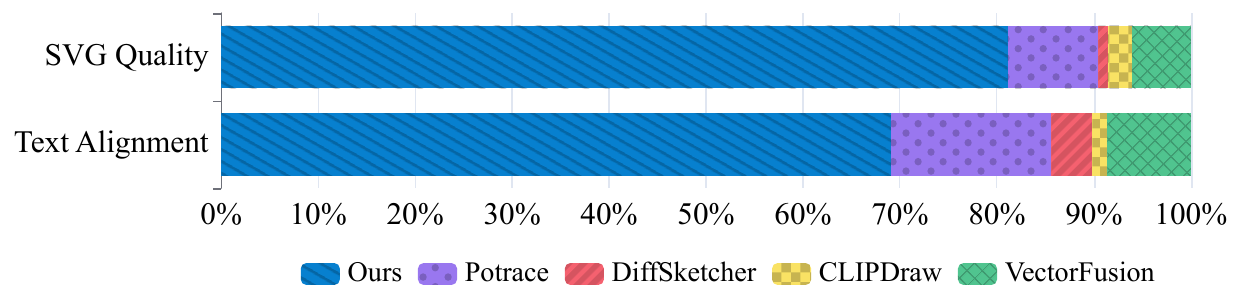}
  \caption{ \label{fig:user_study_bar} User Study. We show the human preferences in \%. }
\end{figure}

\begin{table}[tbp]
  \small
  \caption{ Ablation study on neural path representation learning and text-driven neural path optimization modules. }
  \resizebox{\linewidth}{!}{
    \begin{tabular}{|cl|c|c|c|}
      \hline
      \multicolumn{2}{|c|}{Methods}                                                            & Smooth $\uparrow$ & Simp $\downarrow$ & $Sim_g$ $\uparrow$ \\ \hline
      \multicolumn{1}{|c|}{\multirow{2}{*}{Neural Path Representation (NPR)}} & w/o NPR        & 0.5271            & 40                & 0.9941          \\ \cline{2-5} 
      \multicolumn{1}{|c|}{}                                                  & Sequence       & 0.8116            & 40                & 0.9910          \\ \hline
      \multicolumn{1}{|c|}{\multirow{4}{*}{Neural Path Optimization}}                & w/o VSD    & 0.7896            & 380               & -               \\ \cline{2-5} 
      \multicolumn{1}{|c|}{}                                                  & SDS            & 0.8070            & 40                & 0.9932          \\ \cline{2-5} 
      \multicolumn{1}{|c|}{}                                                  & VSD Only    & 0.8194            & 40                & -               \\ \cline{2-5} 
      \multicolumn{1}{|c|}{}                                                  & w/o Layer-wise & 0.7924            & 40                & 0.9929          \\ \hline
      \multicolumn{2}{|c|}{Ours}                                                               & 0.8012            & 40                & 0.9926          \\ \hline
    \end{tabular}
  }

  \label{tab:ablation_study}
\end{table}

\subsection{User Study}
\label{sec:user_study}

We conducted a perceptual study to evaluate our text-to-vector generation from two perspectives: overall SVG quality and alignment with the text prompt. We randomly selected 20 text prompts from our test set and generated SVGs using both the baseline methods described in \refsec{comparison_with_existing} and our approach.
To gather feedback, we recruited 32 participants (17 males and 15 females) through university mailing lists. The participants had diverse ages, with an average age of 24, and varied levels of design experience.
Each question presented the results of different methods in a random order, and participants were given unlimited time to select the best result among five options for each evaluation metric.
As shown in \reffig{user_study_bar}, our method demonstrated superior performance by achieving the highest preference in both evaluation metrics. Specifically, our method received 81.1\% of votes for overall SVG quality and 69.2\% for text alignment. These results demonstrate the effectiveness of our method in generating high-quality SVGs from text prompts that align more closely with human perception.

\subsection{Ablation Study}
\label{sec:ablation_study}
We conducted ablation studies to validate the effectiveness of key components in our pipeline.

\subsubsection{Ablation on Neural Path Representation Learning}
To illustrate the effectiveness of our neural path representation learning module and the design choice of our dual-branch VAE, we compare them with two baselines.

\begin{figure}[tbp]
  \centering
  \includegraphics[width=1.0\columnwidth]{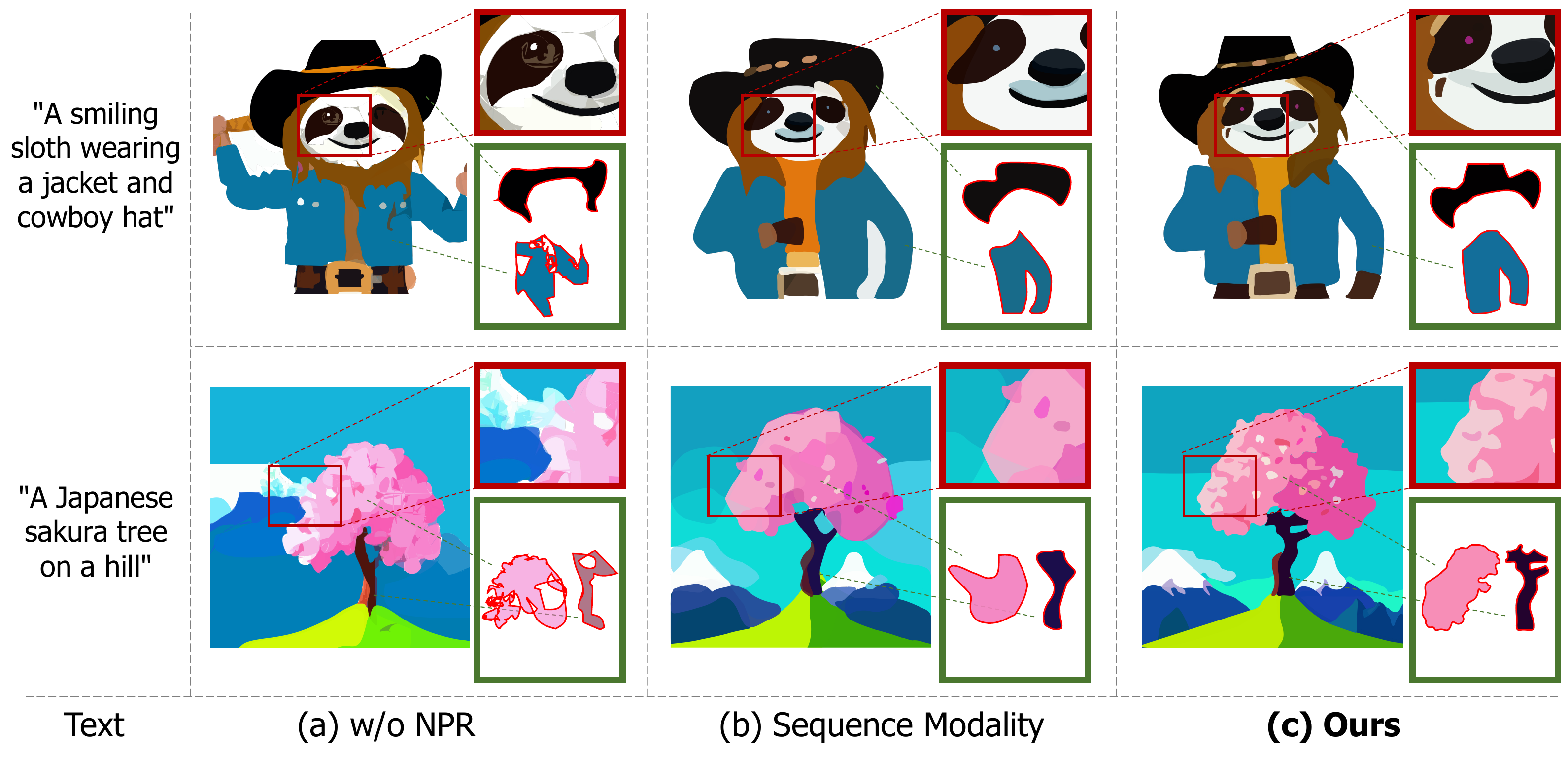}
  \caption{ \label{fig:ablation_npr} Ablation on Neural Path Representation Learning. }
\end{figure}

The first baseline directly optimizes the control points of parametric paths using our text-driven optimization module (\textbf{w/o NPR}). Without the neural path representation, optimized paths often suffer from intersections and abrupt curvature changes. Consequently, this approach generates SVG paths of poor quality characterized by low smoothness score in \reftab{ablation_study}, as shown in \reffig{ablation_npr} (a).

In another baseline, we replace our dual-branch VAE by a sequence VAE with single-modality representation (\textbf{Sequence}). However, relying solely on a sequence representation proves to be insufficient in capturing the visual features necessary for accurately reconstructing the rendered shape. \reffig{ablation_npr} (b) visualizes the resulting challenges in reconstruction. To evaluate the reconstruction accuracy, we compute the image similarity $Sim_g$ between the guidance image and the rendered SVG in RGB space. The results indicate that the sequence VAE achieves a lower $Sim_g$ score compared to our dual-branch VAE (\reftab{ablation_study}).

Our neural path representation proves to be more effective for path optimization, resulting in more accurate paths and better visual quality of the SVGs.

\begin{figure}[tbp]
  \centering
  \includegraphics[width=1.0\columnwidth]{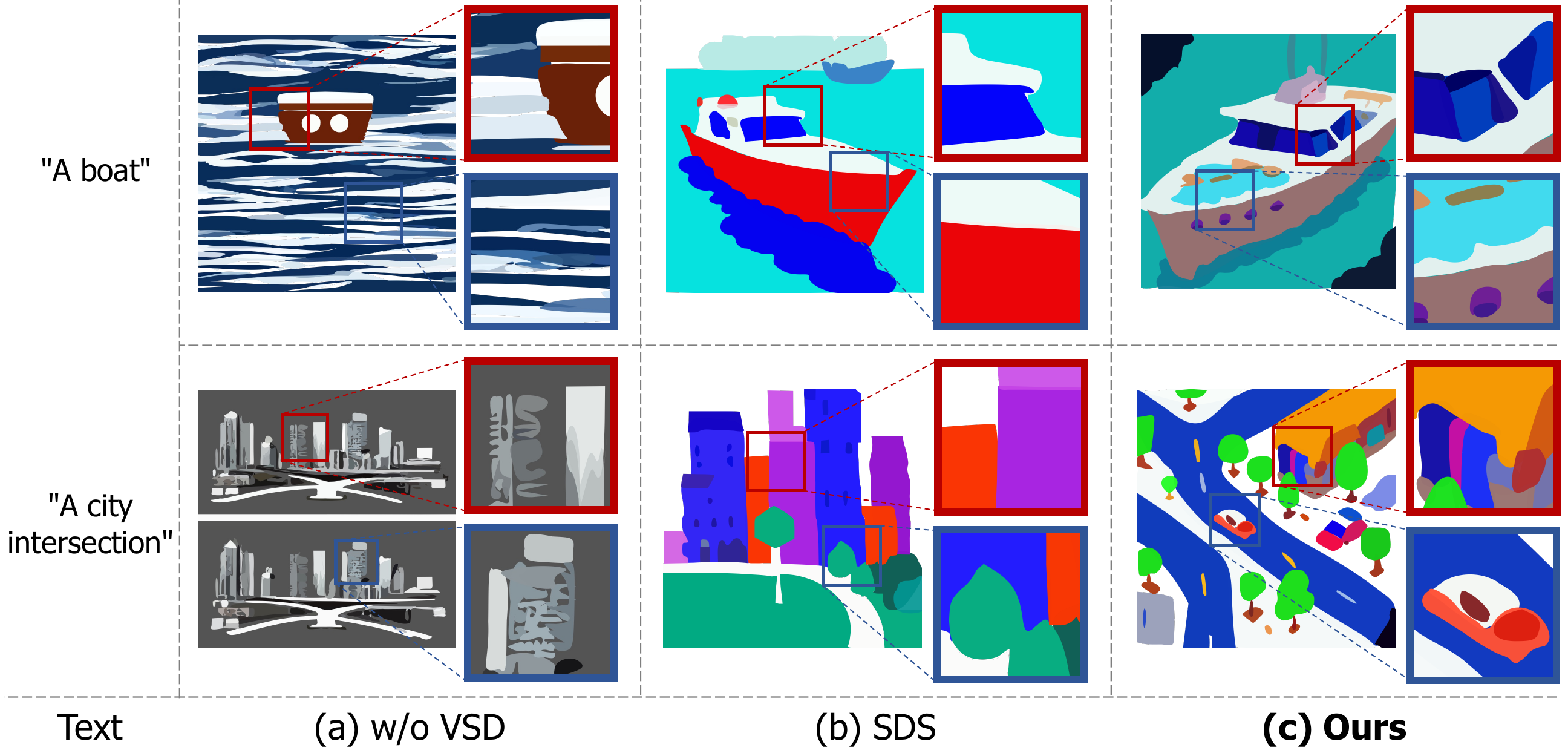}
  \caption{ \label{fig:ablation_tpo_stage1} Ablation on Optimization with Variational Score Distillation. }
\end{figure}

\begin{figure}[tbp]
  \centering
  \includegraphics[width=1.0\columnwidth]{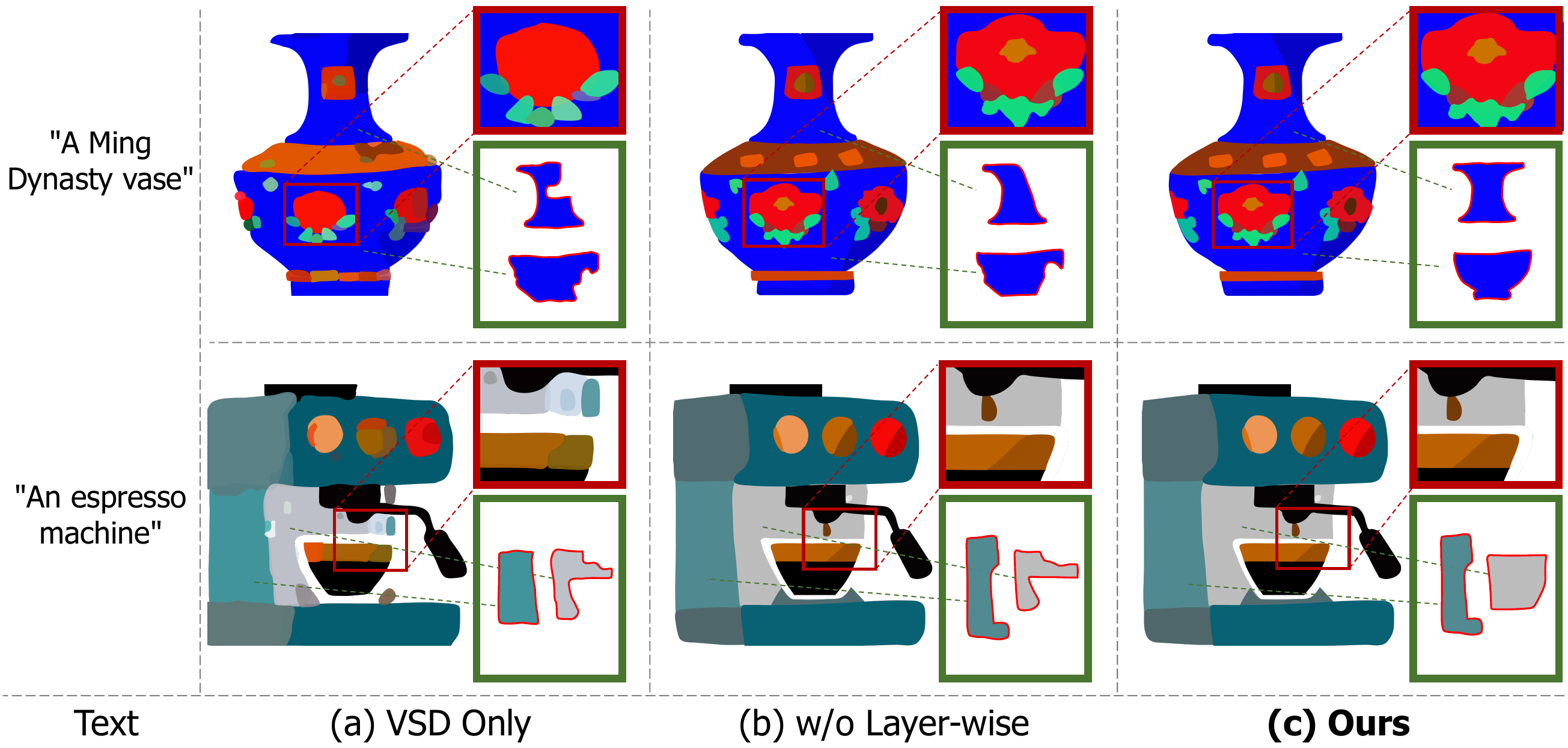}
  \caption{ \label{fig:ablation_tpo_stage2} Ablation on Optimization with Layer-wise Vectorization. }
\end{figure}

\subsubsection{Ablation on Text-driven Neural Path Optimization}
In this subsection, we investigate the effectiveness of our text-driven neural path optimization module.

We first explore the effects of VSD optimization (Stage 1). We generate images using T2I models and employ the layer-wise optimization strategy to transform the raster image into SVGs using our neural path representation (\textbf{w/o VSD}). 
However, T2I models often generate raster images with photographic and realistic styles that do not align with the desired SVG style characterized by smooth geometric shapes and flat colors. The absence of Stage 1 in the optimization process results in SVGs that are overly complex (\reffig{ablation_tpo_stage1}).

We also compare the VSD loss with the SDS loss used in Stage 1. While SDS often leads to oversmoothing and a lack of diversity, VSD generates SVGs with clearer details and higher visual quality, as shown in \reffig{ablation_tpo_stage1}.

We evaluate the impact of optimization with layer-wise vectorization by removing Stage 2 (\textbf{VSD Only}). Without this stage, we observed that SVGs tend to have cluttered and stacked paths. This not only creates visual artifacts but also results in a disorganized layer structure within the SVGs, making them difficult to edit and modify, as shown in \reffig{ablation_tpo_stage2}.

Finally, we compare our layer-wise vectorization strategy with the global vectorization strategy, which involves optimizing all paths together in Stage 2 (\textbf{w/o Layer-wise}). While global vectorization can effectively fit images, it often fails to preserve the topological integrity of SVGs. In contrast, our layer-wise vectorization strategy captures the hierarchical layer structure of SVGs, resulting in clearer and well-structured vector graphics, as shown in \reffig{ablation_tpo_stage2}.

\section{Applications}
\label{sec:application}
We demonstrate the effectiveness of our method with various applications, including SVG generation with adjustable levels of details and different styles, SVG customization, image-to-SVG generation, and SVG animation.

\subsection{SVG Generation with Adjustable Levels of Details}

\begin{figure}[tbp]
  \centering
  \includegraphics[width=1.0\columnwidth]{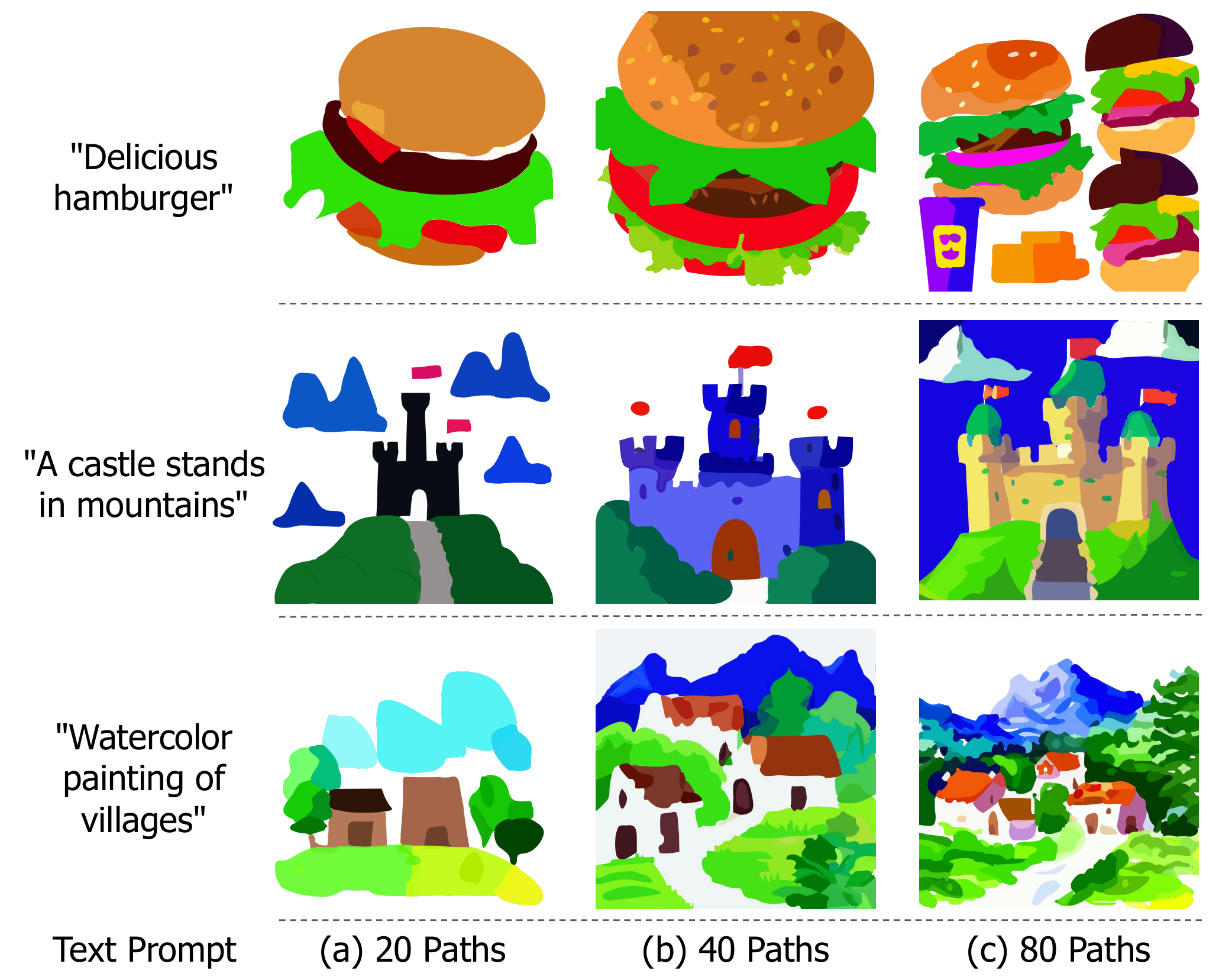}
  \caption{ \label{fig:application_t2v} Controllable text-to-vector with different levels of details. }
\end{figure}

Our method can generate vector graphics with varying levels of abstraction by adjusting the number of paths. \reffig{application_t2v} illustrates the results of generating vector graphics with 20, 40, and 80 paths. Using fewer paths produces SVGs with a simpler and flatter style, while increasing the number of paths adds more detail and complexity.

\subsection{SVG Generation with Different Styles}

\begin{figure}[tbp]
  \centering
  \includegraphics[width=1.0\columnwidth]{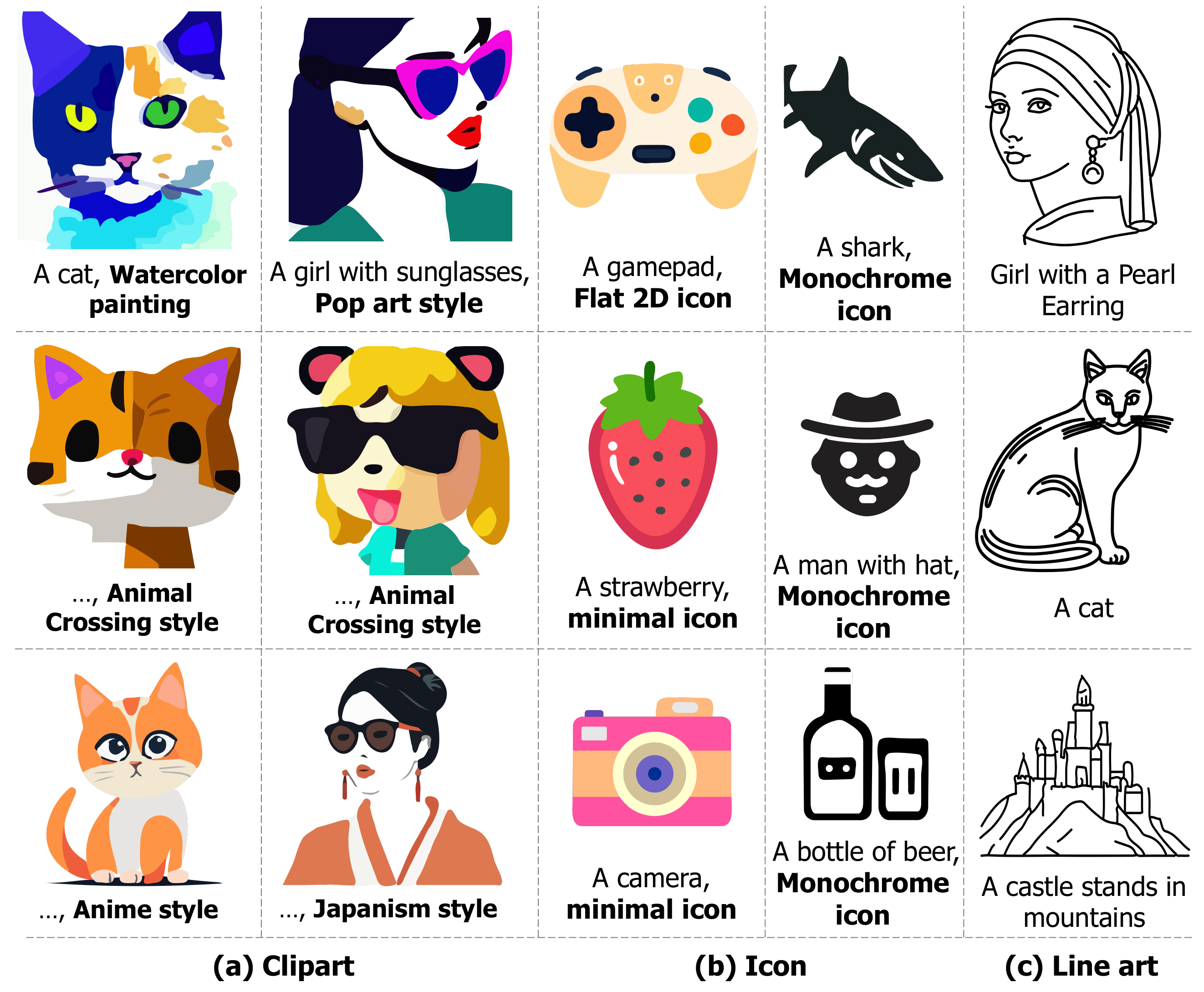}
  \caption{ \label{fig:application_style} Text-to-vector generation with diverse styles. }
\end{figure}

Our method can generate vector graphics with diverse styles by modifying style-related keywords in the text prompts (\eg "watercolor painting" or "Anime style"), or by constraining path parameters such as fill colors and the number of paths. As illustrated in \reffig{application_style}, we append different suffixes to the prompts to produce cliparts and icons that align with the desired aesthetics. To further enhance support for line arts, we fine-tune our path VAE on a dataset of cleaned sketches \cite{yan2020benchmark} featuring open-form paths. Then, by utilizing black color and incorporating "line drawings" into the text prompts, our method can effectively generate line arts.

\subsection{SVG Customization}

\begin{figure}[tbp]
  \centering
  \includegraphics[width=1.0\columnwidth]{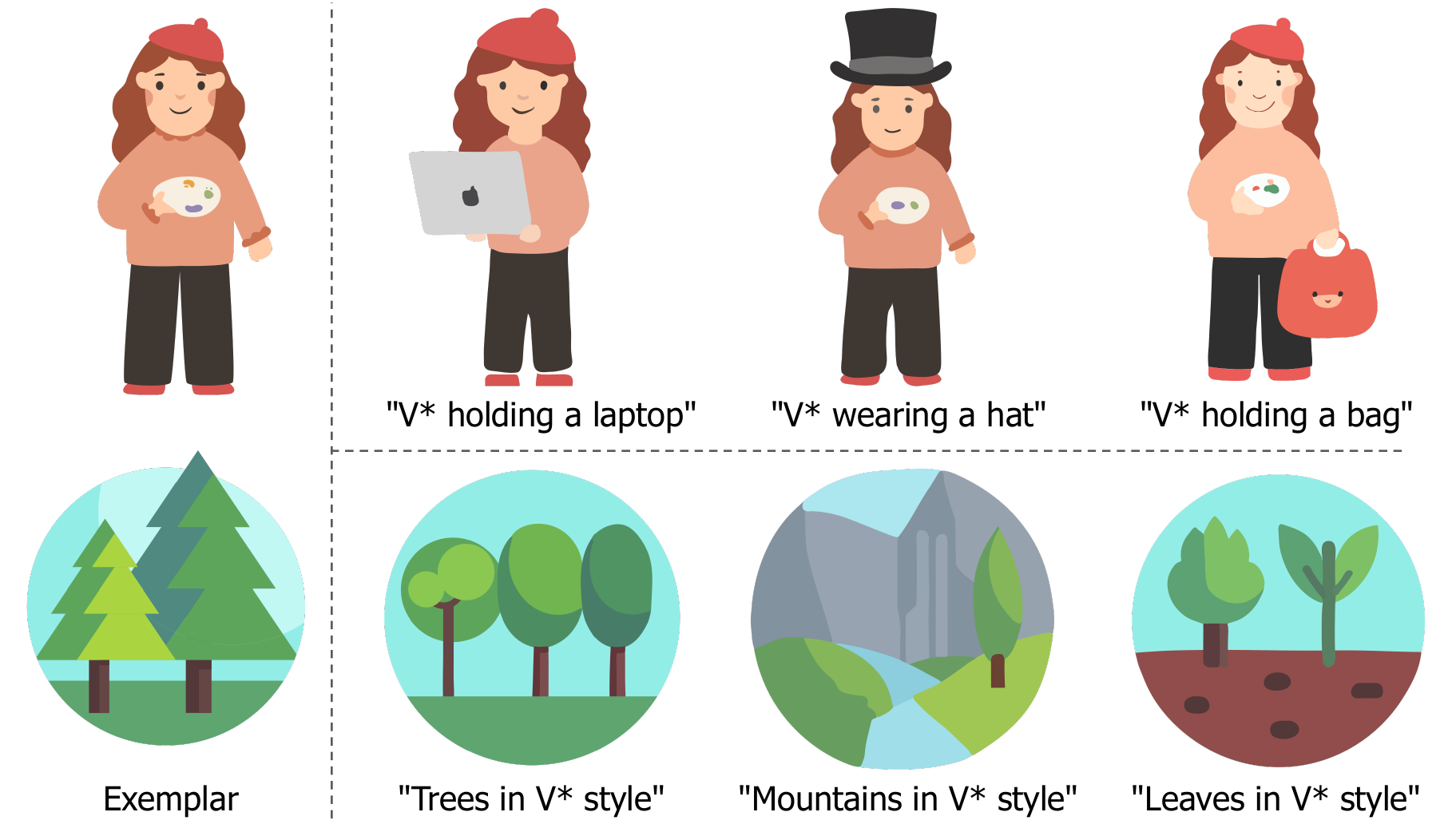}
  \caption{ \label{fig:application_custom} Text-guided SVG customization. Exemplar SVGs: the $1^{st}$ row is from Envato Elements creator \copyright{Telllu}; the $2^{nd}$ row is from \copyright{Freepik}. }
\end{figure}

Given an exemplar SVG, our method can customize the SVG based on text prompts while preserving the visual identity of the exemplar. To achieve this, we follow the approach outlined in \cite{kumari2022multi, zhang2023text} to fine-tune a pre-trained Diffusion model on an exemplar image (rendered from exemplar SVG) and a text prompt containing a special token $V*$. This token learns the concept from the exemplar image. Subsequently, we apply our method to optimize the SVG based on a new prompt, such as "$V*$ holding a laptop" or "Trees in $V*$ style," resulting in a customized SVG that reflects the desired customization in object or style. \reffig{application_custom} showcases the results of customizing the exemplar SVG with different text prompts, demonstrating the flexibility and creativity of our method.

\begin{figure}[tbp]
  \centering
  \includegraphics[width=1.0\columnwidth]{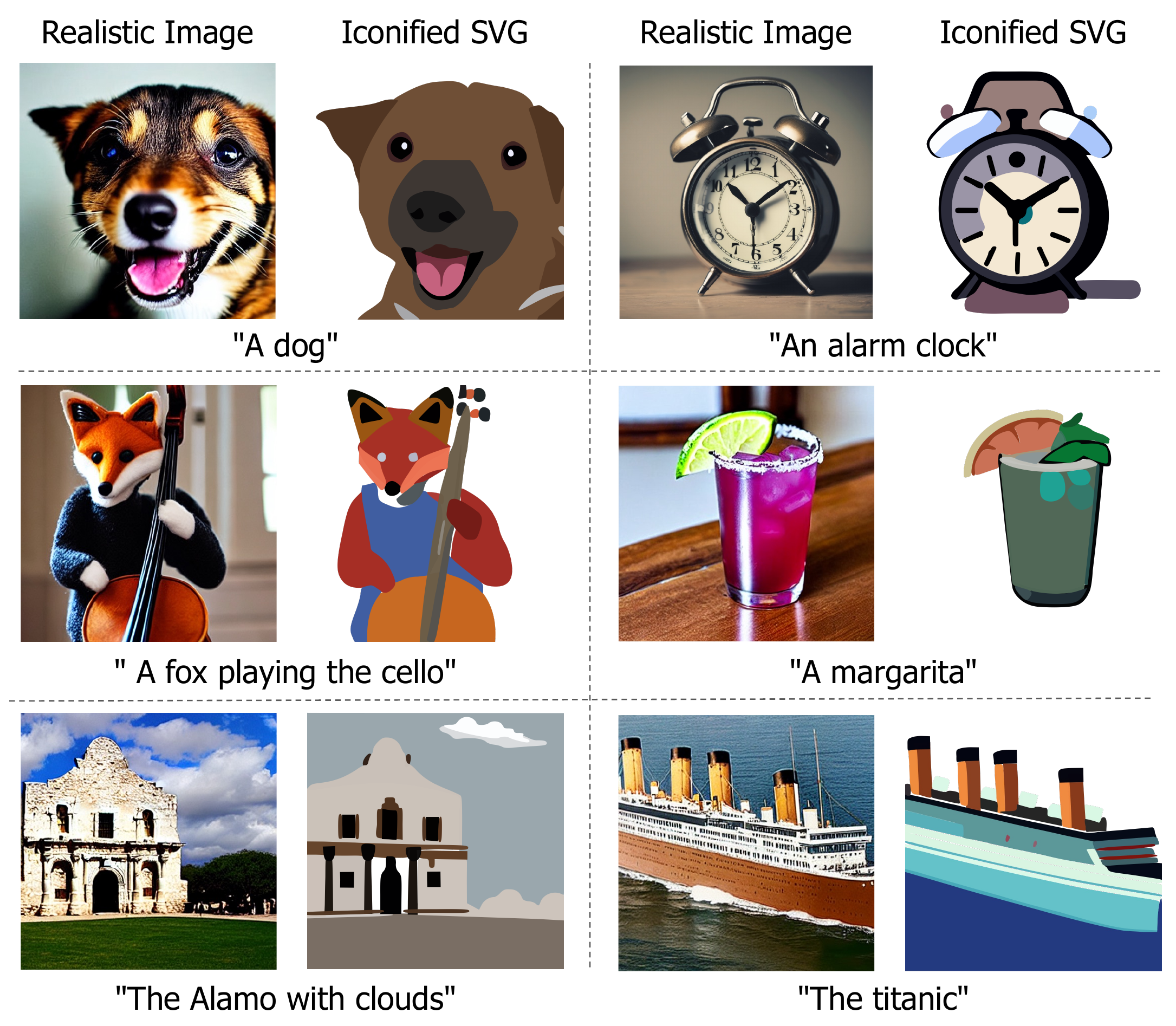}
  \caption{ \label{fig:application_img2icon} Image-to-SVG generation. }
\end{figure}

\subsection{Image-to-SVG Generation}

Our framework enables flexible control beyond text prompts, which is particularly useful for designers seeking inspiration from natural images for SVG-style designs. For example, as shown in \reffig{application_img2icon}, our method can generate vector icons from natural images. This is achieved by integrating the ControlNet \cite{zhang2023adding} into the VSD optimization process, which ensures an optimization direction that respects both the original structure of the image and the input text prompt.

\subsection{SVG Animation}

\begin{figure}[tbp]
 \centering
 \includegraphics[width=1.0\columnwidth]{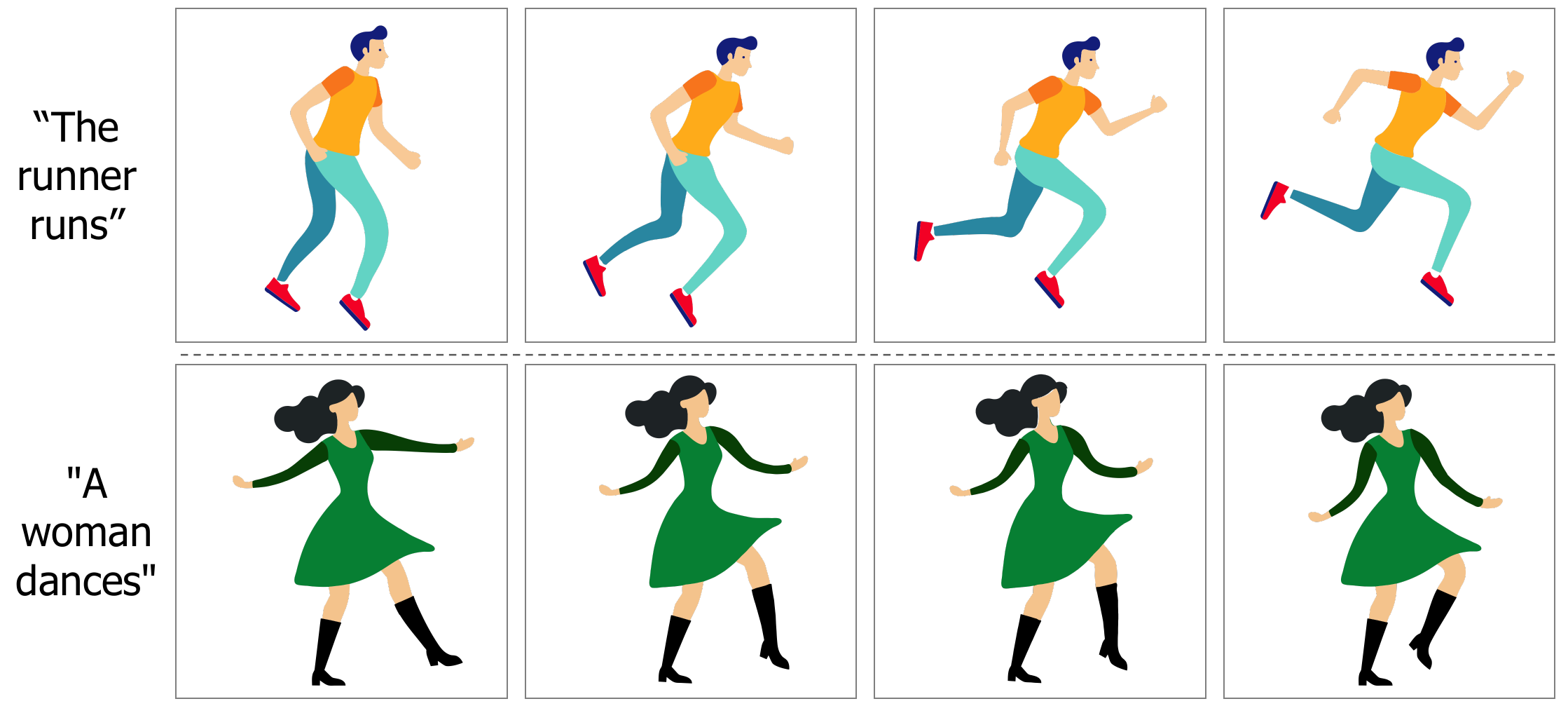}
 \caption{ \label{fig:application_animation} SVG animation aligned with the described motions. }
\end{figure}

Our framework can be extended to SVG animation by animating an initial SVG according to a text prompt describing the desired motion. We first generate a static SVG using our pipeline and then animate it by optimizing a temporal sequence of $m$ paths into $k$ video frames, represented as $Video = \{SVG^1, SVG^2,...,SVG^k\} = \left\{Path_{i}^{j}\right\}_{i \in m}^{j \in k} $, aligning with the motion described in the text prompt. Specifically, we employ a similar two-stage optimization process. In the VSD optimization stage, we leverage a pre-trained text-to-video diffusion model ModelScope \cite{wang2023modelscope} to replace the text-to-image model. In the layer-wise vectorization stage, we generate a guidance video based on the initial SVG sequence \cite{geyer2023tokenflow}, which contains $k$ guidance images. As demonstrated in \reffig{application_animation}, our method animates SVGs with smooth motions, showcasing the effectiveness of our neural path representation.

\begin{figure}[tbp]
  \centering
  \includegraphics[width=1.0\columnwidth]{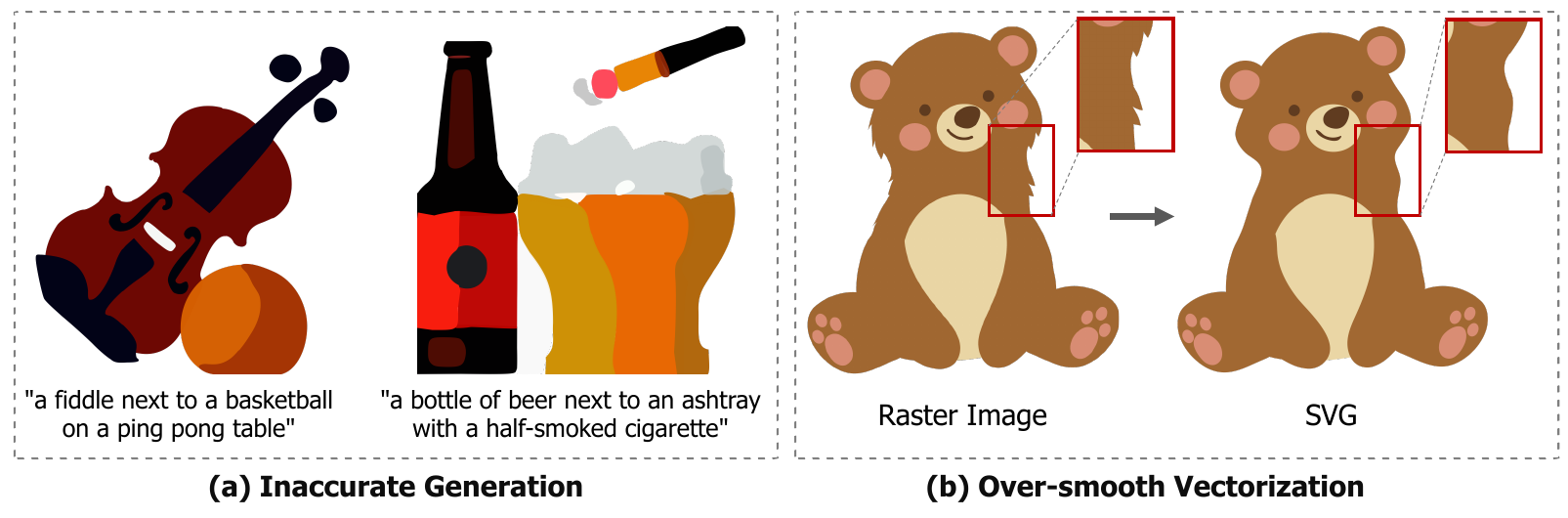}
  \caption{ \label{fig:failure_case} Failure cases. }
\end{figure}

\section{Conclusion}
\label{sec:conclusion}

In this paper, we propose a novel text-to-vector pipeline to generate vector graphics that align with the semantics of given text prompts. Our framework learns a neural path representation within the latent space to capture valid geometric properties of paths. By employing the two-stage text-driven neural path optimization, our method effectively generates vector graphics with desirable path properties and layer-wise structures.

While our method achieves high-quality SVG results, it still has some limitations, as shown in \reffig{failure_case}. First, our method relies on the generative capabilities of the diffusion model; thus, overly detailed text prompts may lead to inaccurate SVG results. For example, as depicted in \reffig{failure_case} (a), semantic elements like the "ping pong table" and "ashtray" are missing.
Second, our method tends to simplify shapes with intricate boundaries into smoother paths, as they exceed the representational capacity of our path latent space. For instance, the detailed edges of the bear's body in \reffig{failure_case} (b) are smoothed out, resulting in the loss of the original complexity. This can be improved by collecting a larger path dataset containing more complex paths, which we leave as future work.
Third, our method, similar to other text-guided SVG optimization methods such as CLIPDraw and VectorFusion, is generally slow due to the iterative optimization process. It takes approximately 13 minutes to optimize 128 paths on an NVIDIA-3090, whereas vectorization with T2I methods can be completed within a few seconds. Despite the current slowness, our neural path representation lays the groundwork for training fast feed-forward T2V networks to replace iterative optimization in future work. This approach could also bring benefits to the generation of graphic layouts, fonts, and CAD models.

\begin{acks}
The work described in this paper was substantially supported by a GRF grant from the Research Grants Council (RGC) of the Hong Kong Special Administrative Region, China [Project No. CityU 11216122].
\end{acks}

\bibliographystyle{ACM-Reference-Format}
\bibliography{sample-acmtog}

\end{document}